\documentclass{article}
\usepackage[preprint]{cpal_2024}

\usepackage{graphicx}
\usepackage{amsmath}
\usepackage{amssymb}
\usepackage{booktabs}

\usepackage{enumitem}
\usepackage{multirow}
\usepackage{bm}
\usepackage{xcolor}

\usepackage[pagebackref,breaklinks,colorlinks]{hyperref}

\usepackage[capitalize]{cleveref}
\crefname{section}{Sec.}{Secs.}
\Crefname{section}{Section}{Sections}
\Crefname{table}{Table}{Tables}
\crefname{table}{Tab.}{Tabs.}

\begin{document}

\title{USDC: Unified Static and Dynamic Compression for Visual Transformer}
\author{
Huan Yuan, Chao Liao, Jianchao Tan\thanks{Corresponding Author.}, ~Peng Yao, Jiyuan Jia, Bin Chen, Chengru Song, Di Zhang
\\
Kuaishou Technology
}

\maketitle

\begin{abstract}
Visual Transformers have achieved great success in almost all vision tasks, such as classification, detection, and so on. However, the model complexity and the inference speed of the visual transformers hinder their deployments in industrial products. Various model compression techniques focus on directly compressing the visual transformers into a smaller one while maintaining the model performance, however, the performance drops dramatically when the compression ratio is large. Furthermore, several dynamic network techniques have also been applied to dynamically compress the visual transformers to obtain input-adaptive efficient sub-structures during the inference stage, which can achieve a better trade-off between the compression ratio and the model performance. The upper bound of memory of dynamic models is not reduced in the practical deployment since the whole original visual transformer model and the additional control gating modules should be loaded onto devices together for inference. To alleviate two disadvantages of two categories of methods, we propose to unify the static compression and dynamic compression techniques jointly to obtain an input-adaptive compressed model, which can further better balance the total compression ratios and the model performances. Moreover, in practical deployment, the batch sizes of the training and inference stage are usually different, which will cause the model inference performance to be worse than the model training performance, which is not touched by all previous dynamic network papers. We propose a sub-group gates augmentation technique to solve this performance drop problem. Extensive experiments demonstrate the superiority of our method on various baseline visual transformers such as DeiT, T2T-ViT, and so on.
\end{abstract}



\section{Introduction}
\label{sec:intro}
Recently, transformer-based models~\cite{vaswani2017attention, yin2022avit} have achieved significant performance improvement in both natural language processing ~\cite{vaswani2017attention,devlin2018bert} and computer vision~\cite{yin2022avit,liu2021swin}. Based on the self-attention mechanism, the ViT model ~\cite{yin2022avit,yuan2021t2tvit} and the subsequent improved ViT model have gradually surpassed CNN-based models in image classification tasks. 
However, the ViT-based model is gradually becoming deeper and more complex,  the huge computing resources and the large parameters of the ViT model will hinder its deployment and application in real-world scenarios.
Therefore, many existing works are devoted to slimming and accelerating ViT models using model compression techniques to reduce the cost of deployment.

The existing compression methods for ViT-based models can be divided into static compression~\cite{chen2021chasing,tang2021patch,yu2021uvc,hou2022multi,pan2021ia} and dynamic compression~\cite{rao2021dynamicvit,xu2022evo,yin2022avit,meng2022adavit,yu2022mia}. 
Static compression methods attempt to permanently and explicitly remove the sub-structure of the network for high efficiency without compromising the performance of the network. 
Because the complexity of the ViT model is proportional to the number of input tokens, Patch Slimming~\cite{tang2021patch} proposed to reduce the computational cost of the ViT models by removing redundant input tokens. 
UVC~\cite{yu2021uvc} proposed to formulate the weight learning, pruning, and layer skipping as a joint optimization problem, thereby compressing the ViT model from multiple dimensions and obtaining minimal performance loss.
Although the static compression method adopts various strategies to reduce the loss of performance while compressing the model, excessively reducing parameters will inevitably destroy the representation ability and robustness of the ViT model.

Compared with static compression, dynamic compression adaptively adjusts the model's parameters and computation graph according to the input during inference which means that the dynamic compression method can accelerate the model without explicitly pruning ~\cite{han2021dynamicsurvey, yang2020resolution,guan2017energy}.
Intuitively, dynamic networks can adaptively use larger computational graphs to process simpler samples and more complex computational graphs to process more difficult samples.
Some existing works have applied dynamic network technology to ViT models.
DynamicViT~\cite{rao2021dynamicvit} reduces redundant tokens hierarchically according to the input tokens' information. 
In DynamicViT~\cite{rao2021dynamicvit}, each encoder layer will decide which tokens need to be used according to the information of the input tokens, instead of using all tokens in each encoder layer by default.
A-ViT~\cite{yin2022avit} early exits encoder layers by dynamic tokens, different tokens dynamically assign different encoder depths. 
AadVit~\cite{meng2022adavit} and MIA Former~\cite{yu2022mia} use the dynamic network as a controller to adaptively skip unnecessary structures according to the input information to dynamic compress the ViT models on multiple dimensions.

\begin{figure*}[!h]
\centering
\includegraphics[width=0.6\textwidth]{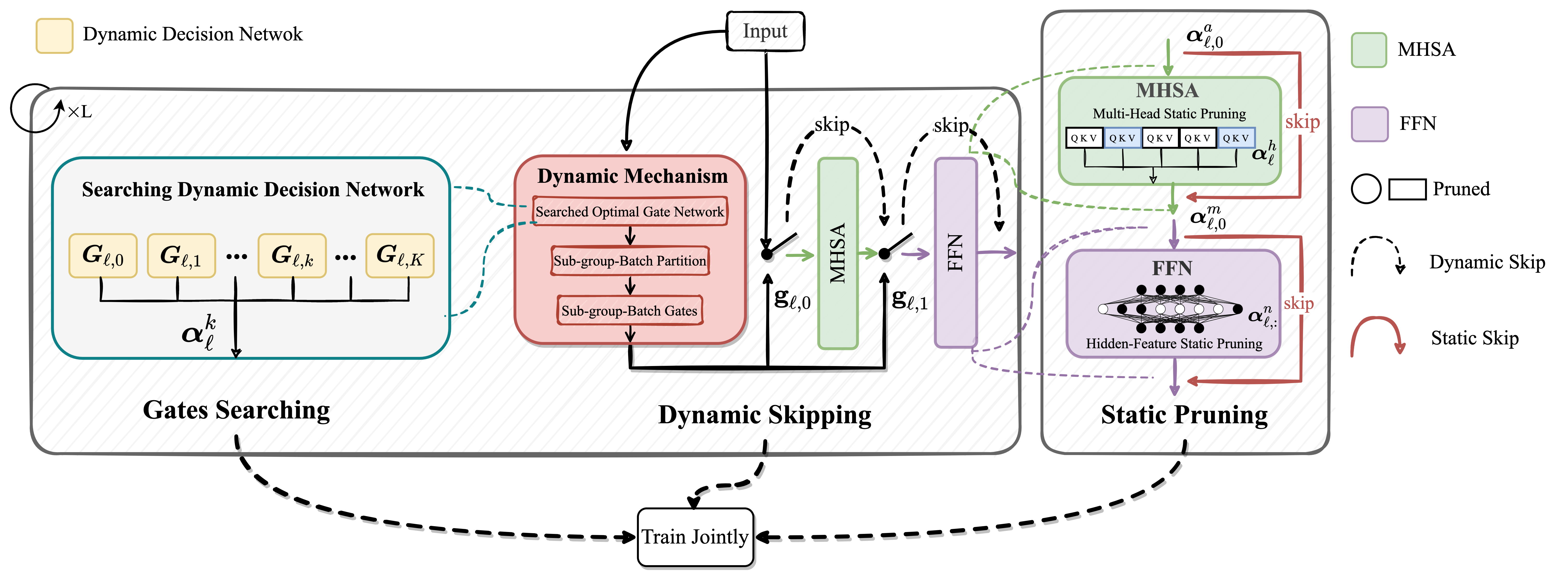}
\caption{ \textbf{An overview of USDC. } We propose to unify the static compression and dynamic compression techniques jointly to obtain an input-adaptive compressed model, which can better balance the total compression ratios and the model performances.}
\label{fig:overview}
\end{figure*}

However, we found that there are several obstacles in the application and deployment of dynamic network technology in real-world scenarios:  
\textbf{First, the dynamic network is not suitable for scenarios that have limited hardware storage}, because it needs to deploy the entire computational graph of the original network to hardware and select sub-computational graphs with different sizes during inference, which means that \textbf{it can reduce the latency but will not reduce the real peak memory}. 
Second, besides the original backbone network, \textbf{dynamic network technology requires an additional gate network as a controller} to decide whether to discard some structures according to the input information, which will \textbf{further increase the size of the computational graph compared with the original model}.
Additionally, the training of dynamic networks is complex, specifically, most dynamic networks are trained at the sample level for flexible decision-making, but the models usually perform inference at the batch level, which means that the gate network only needs to make decisions based on one sample information during training, but needs to make decisions based on all samples within mini-batch during inference. 
\textbf{Generally, we argue that it will lead to the inconsistency of performance between training and inference stages, caused by the different batch sizes between them.
To the best of our knowledge, there is no existing work has considered this problem.}
To address the aforementioned problem, we propose USDC in this work to combine static compression and dynamic compression for vision transformers. 
The overall framework of USDC is shown in Fig.~\ref{fig:overview}, the static compression part can permanently and explicitly reduce the parameters of the ViT model, which can offset the additional parameters brought by the gate network and effectively reduce the maximum storage during deployment.
Furthermore, because encoders at different locations of ViT may have different requirements for dynamic decision networks, we use the neural architecture search (NAS)~\cite{touvron2021deit} technique to automatically select the appropriate gate network for each encoder layer during training, instead of assigning the same gate network to each encoder layer.
In addition, considering that the mini-batch size during deployment is usually smaller than that in training, we propose a novel group-level gates augmentation strategy, which will divide the samples in a mini-batch into groups of different sizes, and the gate network will generate different gates according to the information of each group.
The main contributions are as follows:
\begin{itemize}[itemsep=2pt,topsep=2pt,parsep=1pt]
\item To the best of our knowledge, we are the first to propose a unified dynamic compression and static compression technique for ViT compression. The proposed USDC framework jointly optimizes static compression and dynamic compression in an end-to-end manner, thereby reducing the deployment cost of the model with minimal performance loss.
\item To make a better trade-off between the accuracy and efficiency of the decisions made by the dynamic gates, we propose to use NAS technology to automatically select the appropriate gate network for each layer.
\item A group-level gates augmentation strategy is introduced into the training stages, which improves the performance consistency between the training and inference stages.
\item  On the ImageNet dataset, comprehensive experiments conducted on existing ViT benchmarks show the effectiveness and robustness of our proposed USDC.
\end{itemize}

\section{Related Work}
\subsection{Vision Transformers}
Transformer architectures~\cite{vaswani2017attention, devlin2018bert} is first applied and achieved success in the field of NLP by extracting effective information from sequential data through self-attention mechanism~\cite{vaswani2017attention,devlin2018bert}. Follow transformers~\cite{vaswani2017attention}, the Vision Transformer~\cite{yin2022avit} applies the transformer structure and self-attention layers to the image classification task~\cite{deng2009imagenet}. ViT modes flatten the image into serialized patches and feed them into sequence transformer encoder layers after embedding. Since then, the ViT model continuously has been developed into various variants and obtained the results of SOTA, including DeiT~\cite{touvron2021deit}, CeiT~\cite{touvron2021ceit}, Swin Transformer~\cite{liu2021swin}. T2T-ViT~\cite{yuan2021t2tvit} recursively encodes neighboring important tokens into one token to reduce token length. Swin Transformer~\cite{liu2021swin} uses shifted windows to achieve flexibility at various scales. Moreover, the application of ViT models is extended to more computer vision tasks, such as semantic segmentation~\cite{wang2021max, valanarasu2021medical}, object detection~\cite{carion2020end}, video understanding~\cite{arnab2021vivit, liu2022video, fan2021multiscale}.

\subsection{Dynamic Networks}
Dynamic networks adapt model architectures and parameters according to input during inference~\cite{han2021dynamicsurvey}. For a simpler sample, only a smaller computational graph and fewer parameters are needed to compute. For more complex samples, the dynamic network adaptively uses more parameters and a larger computational graph to compute. Dynamic network not only improves the efficiency but also ensures the representation ability of the model. Recently, dynamic network structures have been widely used in deep learning~\cite{han2021dynamicsurvey,wang2018skipnet,guo2019dynamic}. ~\cite{kaya2019shallow,guan2017energy,bolukbasi2017adaptive, jie2019anytime, yang2020resolution} accelerates the inference of the networks by early exits. ~\cite{yang2020resolution} dynamic routing in different sub-networks with different resolutions. However, the later layers of networks usually have important semantic features and are unsuitable for early exit strategies. ~\cite{wang2018skipnet, figurnov2017spatially} skips the middle layers or blocks by layer skipping mechanism to obtain the dynamic depth of the DNN models. ~\cite{figurnov2017spatially} adaptively stopping the inference of residual networks~\cite{he2016deep} by halting scores to skip residual blocks dynamically. ~\cite{wang2018skipnet} learns whether the current residual block needs to be skipped through the dynamic gate network. Learnable dynamic gate networks are widely used to control the skipping of dynamic depth networks. Dynamic gates networks usually are set as a simple linear layer~\cite{meng2022adavit} or serial multiple linear layers~\cite{wang2018skipnet, guo2019dynamic, veit2018convolutional} with pooling operator and BatchNorm layer~\cite{ioffe2015batch}. There are some works that applied dynamic network technology to the ViT models. ~\cite{rao2021dynamicvit,yin2022avit,wang2021not} apply the dynamic mask on the token to improve the efficiency of the ViT model, and ~\cite{yu2022mia, meng2022adavit} implement the dynamic network on ViT models at multiple dimensions, including heads, tokens, blocks.

\subsection{Static Model Compression}
The increasing depth and size of deep neural networks have brought great challenges to the deployment of deep learning. The compression method of static pruning can explicitly downsize the model and has always been used to reduce the number of parameters and computation of DNN models~\cite{guo2021gdp, yu2021uvc, shen2020umec}.
According to the granularity of pruning, static compression can be divided into two types: unstructured pruning~\cite{xiao2019autoprune, lee2018snip} and structured pruning~\cite{lin2018accelerating, han2015learning,yu2021uvc, guo2021gdp, he2019filter,molchanov2019importance}. Unstructured pruning methods sparsify the elementwise weight values of the model, which can reduce the storage capacity of the model, but it is difficult for most of the current hardware to support the reasoning of this sparse compressed model. Structured pruning usually takes channels~\cite{molchanov2019importance, guo2021gdp, Ding_2021_ICCV}, layers, blocks~\cite{yu2021uvc}, etc. as a unit, and structurally removes the entire unit. Structured pruning is widely used in DNN because it can slim down the model well and is hardware-friendly.
Structured pruning, through the basic pruning unit, is one or more Channels of filters or weight matrices, can slim down the model, and also has hardware-friendly features. The importance or magnitude of each channel of the conv-layer can be associated with the batch norm layer\cite{molchanov2019importance, he2017channel}. ~\cite{Liu2017learning} use the corresponding scaling factor in the batch norm layer after a channel as the importance of the channel.
Recently, there have been more and more static structured pruning methods for ViT models. 
Patch slimming~\cite{tang2021patch} reduces the input of the model by removing redundant input tokens. UVC~\cite{yu2021uvc} optimizes the structure of  ViT models through the dual optimizer method to obtain a lightweight model. ~\cite{hou2022multi} apply multi-dimensional compression on the  ViT models through guided Gaussian process search.
The complexity of the ViT model makes it have better performance, but the high storage space and computing resource consumption are important reasons for it to be difficult to effectively apply on various hardware platforms~\cite{zhu2021visual, liu2021posttraining}.

\section{Approach}

\subsection{ViT Preliminaries}
Vision Transformers~\cite{dosovitskiy2020vit, touvron2021deit, yuan2021t2tvit} split the input images into raw patches and embed these fixed-size patches into tokens by patch embeddings and positional embeddings, and then those tokens will be feed into sequential transformer encoder layers. Similar to transformers~\cite{vaswani2017attention,devlin2018bert} in NLP, the encoder layers of ViT models usually consist of multi-head self-attention (MHSA) block and feed-forward network (FFN) block. MHSA block combines the output self-attentions information of heads, every single head $\operatorname{head}_{i,l}$ project all input tokens $\boldsymbol{Z}_{\ell}$ into three matrices by three linear layers, named $\boldsymbol{Q}_{i, \ell}$, $\boldsymbol{K}_{i, \ell}$ and $\boldsymbol{V}_{i, \ell}$, where $l$ and $i$ denote the $l_{th}$ encoder of ViT and $i_{th}$ head of MHSA. 
\begin{equation}
\small
\begin{aligned} 
\label{eq:qkv}
    &\boldsymbol{Q}_{i, \ell}=\boldsymbol{Z}_{\ell} \boldsymbol{W}_{i, \ell}^Q \\
    & \boldsymbol{K}_{i, \ell}=\boldsymbol{Z}_{\ell} \boldsymbol{W}_{i, \ell}^K \\
    & \boldsymbol{V}_{i, \ell}=\boldsymbol{Z}_{\ell} \boldsymbol{W}_{i, \ell}^V 
\end{aligned}
\end{equation}
where $\boldsymbol{W}^{Q}$, $\boldsymbol{W}^{K}$ and $\boldsymbol{W}^{V}$ denote the trainable parameters of three linear layer.
Then these three matrices will be used to calculate the  self-attention as shown in Eq.~\ref{eq:attn}:
\begin{align} \label{eq:attn}
\small
    \operatorname{Attn}(\boldsymbol{Q}, \boldsymbol{K}, \boldsymbol{V})=\operatorname{Softmax}\left(\frac{\boldsymbol{Q} \boldsymbol{K}^{\top}}{\sqrt{d}}\right) \boldsymbol{V}
\end{align}%
Finally, MHSA concatenates the result from every head and projects them as the output by linear layer. FFN block takes the output token of MHSA as input and then projects it by two-layer linear: $\{\boldsymbol{W}_{ \ell}^{I},\boldsymbol{W}_{\ell}^{O}\}$. For simplicity, in all equations, the bias of the linear layers is not mentioned.
\begin{equation}
\begin{small} 
\begin{aligned} 
\label{eq:headffn1}
    &\operatorname{head}_{i, \ell} =\operatorname{Attn}\left(\boldsymbol{Z}_{\ell} \boldsymbol{W}_{i, \ell}^Q, \boldsymbol{Z}_{\ell} \boldsymbol{W}_{i, \ell}^K, \boldsymbol{Z}_{\ell} \boldsymbol{W}_{i, \ell}^V\right) \\
    &\operatorname{MHSA}\left(\boldsymbol{Z}_{\ell}\right) =\operatorname { Concat }\left(\operatorname{head}_{1, \ell}, \ldots, \operatorname{head}_{H, \ell}\right) \boldsymbol{W}_{\ell}^O \\
    &\operatorname{FFN}\left(\boldsymbol{Z}_{\ell}^{\prime} \right) = LN\left(\boldsymbol{Z}_{\ell}^{\prime} \right)\boldsymbol{W}_{ \ell}^{I}\boldsymbol{W}_{\ell}^{O}\\
\end{aligned}
\end{small} 
\end{equation}
The LayerNorm~\cite{ba2016layernorm} and residual connections are applied on both MHSA and FFN blocks, then the output of $\ell_{th}$ layer is defined as:
\begin{equation}
\small
\begin{aligned}
\label{eq:block1}
    &\boldsymbol{Z}_{\ell}^{\prime} = \operatorname{MHSA}\left(\operatorname{LN}\left(\boldsymbol{Z}_{\ell-1}\right)\right)+\boldsymbol{Z}_{\ell-1}, \\
    &\boldsymbol{Z}_{\ell}=\operatorname{FFN}\left(\operatorname{LN}\left(\boldsymbol{Z}_{\ell}^{\prime}\right)\right)+\boldsymbol{Z}_{\ell}^{\prime},
\end{aligned}
\end{equation}
The class token of the last transformer encoder layer will be fed into a linear layer, and the output of this linear layer will be used for final classification prediction.
\subsection{Dynamic compression}
\label{sec:dynamic}

Dynamic compression for ViT will choose different computational paths during inference according to the input of each encoder layer. Dynamic networks adaptively use smaller computational graphs to process simple samples and more complex computational graphs to process difficult samples. 

Unlike previous work~\cite{rao2021dynamicvit, yin2022avit}, we do not apply dynamic compression at the token level or width level (heads, channels), because dynamic compression at the token level and width level needs an extra mask or slice operation, which usually introduces extra computation and add time-consuming during inference.  
We want to implement a dynamic compression method that can really speed up ViT models on hardware during real deployment.
For this purpose, USDC applies dynamic compression on the MHSA block and FFN block of ViT models.  Especially, USDC implements a dynamic compression mechanism on all transformer layers of ViT models including the first encoder layer.
The block-level dynamic compression can actually skip the block that the dynamic decision network thinks the current input does not need to execute during the inference. 
In practice, when the dynamic decision network chooses to skip the MHSA or FFN block, the LayerNorm and the residual connections in them will also be skipped. For $\ell_{th}$ encoder layer, the block in Eq.~\ref{eq:block1} becomes:
\begin{equation}
\begin{small}
\begin{aligned}
&\boldsymbol{Z}_{\ell}^{\prime} = \operatorname{MHSA}\left(\operatorname{LN}\left(\boldsymbol{Z}_{\ell-1}\right)\right)+\boldsymbol{Z}_{\ell-1}, \\
&\color{gray} \boldsymbol{Z}_{\ell}^{\prime\prime} =\textbf{g}_{\ell,0}\boldsymbol{Z}_{\ell}^{\prime}  + \left(1-\textbf{g}_{\ell,0}\right)\boldsymbol{Z}_{\ell-1}, \\
&\boldsymbol{Z}_{\ell}^{\prime\prime\prime}=\operatorname{FFN}\left(\operatorname{LN}\left(\boldsymbol{Z}_{\ell}^{\prime\prime}\right)\right)+\boldsymbol{Z}_{\ell}^{\prime\prime} \\
&\color{gray} \boldsymbol{Z}_{\ell}=\textbf{g}_{\ell,1}\boldsymbol{Z}_{\ell}^{\prime\prime\prime}+ \left(1-\textbf{g}_{\ell,1}\right)\boldsymbol{Z}_{\ell}^{\prime\prime},
\end{aligned}
\end{small}
\label{eq:block2}
\end{equation}
Where $\textbf{g}_{\ell,:} \in \mathbf{R}^{1 \times 2}$ are the output of  dynamic decision network  at $\ell_{th}$ encoder layer. To avoid unnecessary calculations, we only apply one dynamic decision network for each encoder transformer layer. The dynamic decision network at $\ell_{th}$ encoder layer will output two gates, the gate at  $\textbf{g}_{\ell,0}$ control the skipping of MHSA at $\ell_{th}$ encoder layer, and the gate $\textbf{g}_{\ell,1}$  control the skipping of FFN. 
During training,  the gates $\textbf{g}_{\ell,:} $ would be relaxed with a hard Gumbel-Softmax trick to keep it differentiable. 

\subsubsection{Dynamic Decision  Network Search}

Dynamic decision networks are widely used to control the skipping of dynamic networks. In previous work, the dynamic decision network is usually set as a simple one linear layer~\cite{meng2022adavit} or serial multiple linear layers with pooling operator and activation functions~\cite{wang2018skipnet, guo2019dynamic, veit2018convolutional}. 
Dynamic decision networks with simple architecture may lead to insufficient learning of skipping redundant layers and make inaccurate decisions of dynamic compression.
However, a complex dynamic decision network will add too many computational resources, which is contrary to the goal of compression. Each encoder layer may have different requirements for a dynamic decision network.
We can use an automated architecture searching method to find the optimal dynamic decision network, instead of using the same decision network architecture for each encoder layer in the whole network such as previous work~\cite{meng2022adavit,veit2018convolutional,guo2019dynamic}. 
Therefore, we construct the dynamic decision networks searching space, which contains different complexities, normalization layers (BatchNorm~\cite{ioffe2015batch}, LayerNorm~\cite{ba2016layernorm}), and activation functions (ReLU, GeLU). 
USDC applies differentiable architecture search technology~\cite{liu2018darts} on searching gate networks for each transformer layer. 

As shown in Fig.~\ref{fig:overview}, for the $\ell_{th}$ transformer encoder layer, the searching space consists of $K$ dynamic decision networks $\{\boldsymbol{G}_{\ell,0}\, \boldsymbol{G}_{\ell,1} ,..., \boldsymbol{G}_{\ell,K}\}$ as follows:
\begin{itemize}[itemsep=2pt,topsep=2pt,parsep=1pt]
\item $\boldsymbol{G}_{\ell,0}$ :  two FC layer with LinearNorm and ReLU.
\item $\boldsymbol{G}_{\ell,1}$ :  two FC layer with BatchNorm and ReLU.
\item $\boldsymbol{G}_{\ell,2}$ :  two FC layer with LinearNorm and GeLU.
\item $\boldsymbol{G}_{\ell,3}$ :  one FC layer.
\item $\boldsymbol{G}_{\ell,4}$ :  two convolutional layer with $1\times1$ kernel size and BatchNorm and ReLU.
\item $\boldsymbol{G}_{\ell,5}$ :  two convolutional layer with $1\times1$ kernel size BatchNorm and GeLU.
\item $\boldsymbol{G}_{\ell,6}$ :  one convolutional layer with $1\times1$ kernel size .
\end{itemize}

For each dynamic decision network, we set a learnable architecture parameter $\alpha_{\ell,k}^g$ to represent the importance of the corresponding $k_{th}$ dynamic decision network. During the training of the ViT, the architecture parameter will be updated by the training loss. The gates $\textbf{g}_{\ell}$  of  $\ell_{th}$ block is: 

\begin{align}
\small
&\textbf{g}_{\ell} = \sum_{k \in K} \alpha_{\ell,k}^g\boldsymbol{G}_{\ell,k}\left(\boldsymbol{Z_{\ell-1}}\right)
\end{align}

The $\alpha_{\ell,k}^g$ is the Gumbel-Softmax~\cite{maddison2016gumbel} function, which aims to relax the discrete search space into a differentiable and continuous search space. After the searching, the candidate gate networks corresponding to the largest value in $\alpha_{\ell,:}^g$ will be selected as the dynamic decision networks of the current transformer encoder layer.

\subsubsection{Augmentation for Dynamic Gates}


In addition to using different gate networks for encoders at different locations in the network, whether the samples within a mini-batch ought to use the same gate at any location of ViT is also a heuristic question.
Some previous work directly adopts the \textbf{sample-level} strategy~\cite{wang2018skipnet, meng2022adavit,veit2018convolutional,guo2019dynamic} to generate a unique gate for each sample within a mini-batch, which may have better performance but hardware-unfriendly. During the inference, the hardware processes samples within a mini-batch in a parallel manner, and the sample-level strategy will destroy the calculation parallelism.

The \textbf{batch-level} strategy generates one same gate for all samples within a mini-batch, which is hardware-friendly. However, when the batch size is large, the complexity of samples within a mini-batch could be various, and it is difficult for the batch-level dynamic network to achieve good performance. 
In addition, in real-world scenarios, training and inference often use different batch sizes (e.g. Using large mini-batch size for training and small for inference), so batch-level strategies may lead to inconsistent performance between training and inference stages.

\begin{figure}[!t]
\centering
\includegraphics[width=0.4\textwidth]{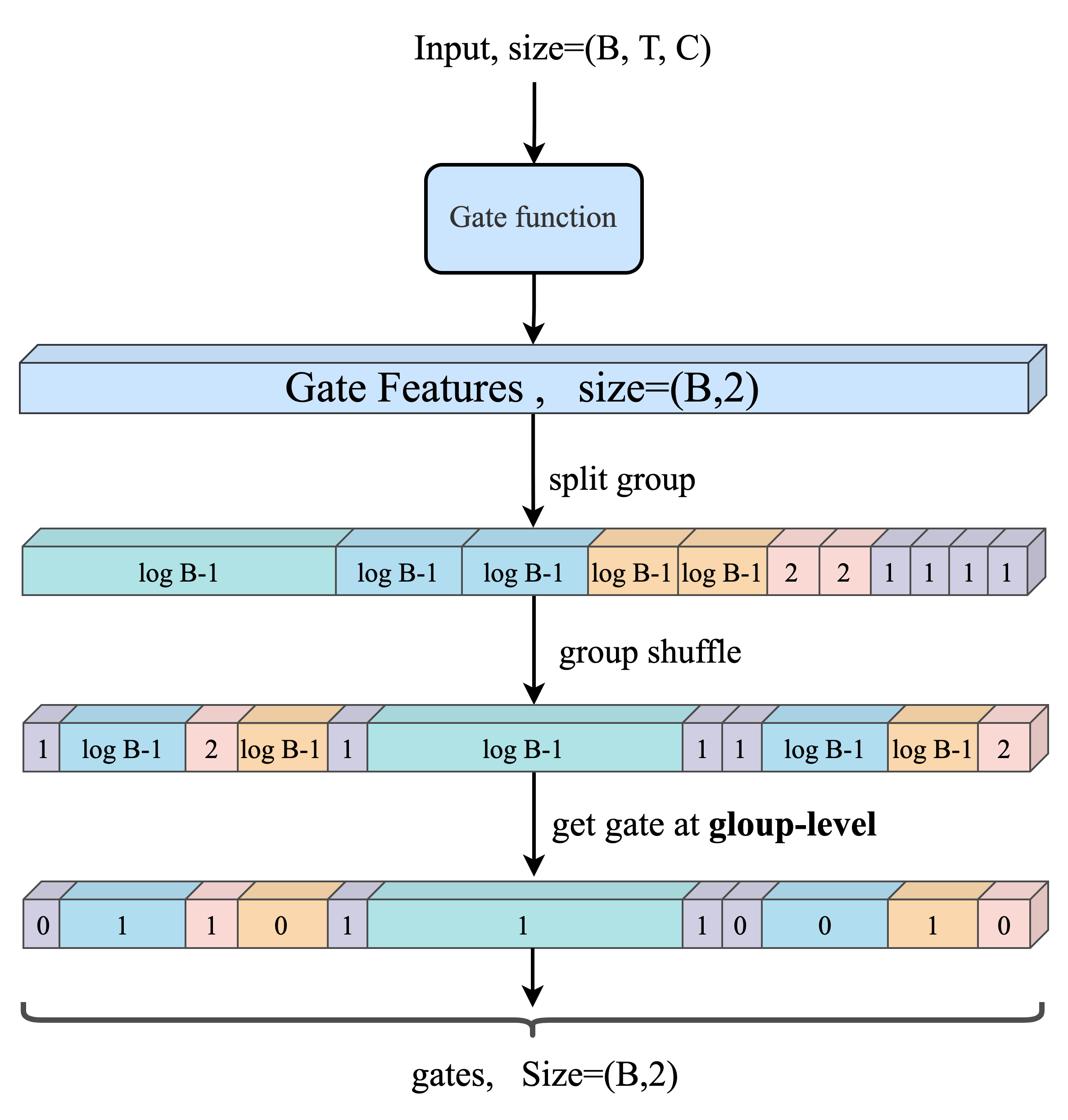}
\caption{Group-level dynamic gates augmentation.}
\label{fig:group}
\end{figure}

To address the aforementioned problem, we propose a \textbf{group-level} strategy to augment dynamic gates. As shown in Fig.~\ref{fig:group}, $B$ is the mini-batch size during training, we recursively split the gates features into the different groups by a logarithm of 2.
After randomly shuffling these groups, the gate of each group will be achieved by calculating the average of the gate features within the same group, the gate values of each sample inside the same group are consistent. 


In summary, the group-level strategy can effectively randomly use the sample-level strategy and the batch-level strategy of different mini-batch sizes during the training process, which enables the gate network to learn to make accurate decisions based on the information of all the samples under various mini-batch sizes.
We can use a group-level strategy during the training of dynamic networks, to reduce the performance drop at the inference stage, which is caused by using different batch sizes between the training and testing stages, then we can use a batch-level strategy during the inference stage to obtain hardware-friendly acceleration without a performance drop.

\subsection{Static compression}
Static compression explicitly prunes some structure of the model to obtain a lightweight model. For ViT transformer encoder layers, we apply structured static compression on the blocks, the number of heads, and the dimension of the channel, the blocks include MHSA and FFN. 

We combine dynamic and static compression for efficient ViT models. For the $\ell_{th}$ transformer encoder layer in a ViT model,  we set learnable parameters $\{\boldsymbol{\alpha}_{\ell}^a, \boldsymbol{\alpha}_{\ell}^m, \boldsymbol{\alpha}_{\ell}^h, \boldsymbol{\alpha}_{\ell}^n\}$
 as static compression pruning parameters. During training,  all of the static compression parameters  $\{\boldsymbol{\alpha}_{\ell}^a, \boldsymbol{\alpha}_{\ell}^m, \boldsymbol{\alpha}_{\ell}^h, \boldsymbol{\alpha}_{\ell}^n\}$ would be relaxed with soft Gumbel-Softmax trick to keep it differentiable.  $\{\boldsymbol{\alpha}_{\ell}^h \in \mathbf{R}^{H \times 2}\}$ are the static compression parameters for multi-head attention in transformer layers, the {H} is the head number.  $\{\boldsymbol{\alpha}_{\ell}^n \in \mathbf{R}^{N \times 2}\}$ are for hidden layer in FFN block, the $N$ is the hidden layer's  channel  dimension. Similar to dynamic gate network searching, we use a differentiable architecture method~\cite{liu2018darts} to optimize static compression parameters. Under static compression training, the MHSA and the FFN in ViT models defined in Eq.~\ref{eq:headffn1} would be calculated as follows:
\begin{equation}
\begin{small}
\begin{aligned} 
\label{eq:head}
    &\operatorname{head^{\prime}}_{\ell, i} =\boldsymbol{\alpha}_{\ell,i, 0}^h\operatorname{Attn}\left(\boldsymbol{Q}_{\ell,i}, \boldsymbol{K}_{\ell,i}, \boldsymbol{V}_{\ell,i}\right) \\
    &\operatorname{MHSA^{\prime}}\left(\boldsymbol{Z}_{\ell}\right) =\operatorname { Concat }\left(\operatorname{head^{\prime}}_{\ell,1}, \ldots, \operatorname{head^{\prime}}_{\ell, H}\right) \boldsymbol{W}_{\ell}^O \\
        &\operatorname{FFN^{\prime}}\left(\boldsymbol{Z}_{\ell}^{\prime} \right) = LN\left(\boldsymbol{Z}_{\ell}^{\prime} \right)\boldsymbol{W}_{\ell}^{I}\boldsymbol{\alpha}^n_{\ell,:,0}\boldsymbol{W}_{\ell}^{O}
\end{aligned}%
\end{small}
\end{equation}
\noindent Meanwhile, the static compression parameters $\{\boldsymbol{\alpha}_{\ell}^a \in \mathbf{R}^{1 \times 2}\}$ are for MHSA block and  $\{\boldsymbol{\alpha}_{\ell}^f \in \mathbf{R}^{1 \times 2}\}$  are for FFN block in $\ell_{th}$ transformer encoder layer. We optimize the dynamic compression and static compression jointly, therefore, the dynamic gates $\textbf{g}$ and static compression parameters $\boldsymbol{\alpha}$ will be calculated and trained together, the block defined in Eq.~\ref{eq:block2} becomes:
\begin{equation}
\begin{small}
\begin{aligned}
&\boldsymbol{Z}_{\ell}^{\prime} = \operatorname{MHSA^{\prime}}\left(\operatorname{LN}\left(\boldsymbol{Z}_{\ell-1}\right)\right)+\boldsymbol{Z}_{\ell-1}, \\
&\color{gray} \boldsymbol{Z}_{\ell}^{\prime\prime} =\textbf{g}_{\ell,0}\boldsymbol{\alpha}_{\ell,0}^a\boldsymbol{Z}_{\ell}^{\prime}  + \left(1-\textbf{g}_{\ell,0}\right)\boldsymbol{\alpha}_{\ell,1}^a\boldsymbol{Z}_{\ell-1}, \\
&\boldsymbol{Z}_{\ell}^{\prime\prime\prime}=\operatorname{FFN^{\prime}}\left(\operatorname{LN}\left(\boldsymbol{Z}_{\ell}^{\prime\prime}\right)\right) +\boldsymbol{Z}_{\ell}^{\prime\prime} \\
&\color{gray} \boldsymbol{Z}_{\ell}=\textbf{g}_{\ell,1}\boldsymbol{\alpha}_{\ell,0}^m\boldsymbol{Z}_{\ell}^{\prime\prime\prime}+ \left(1-\textbf{g}_{\ell,1}\right)\boldsymbol{\alpha}_{\ell,1}^m\boldsymbol{Z}_{\ell}^{\prime\prime},
\end{aligned}
\end{small}
\end{equation}

\noindent After static compression, the structure (head, channel, and block)  with static parameters meet the conditions of  $ (\boldsymbol{\alpha}_{:,0} < \boldsymbol{\alpha}_{:,1})$ would be pruned. The retained structure will continue to be optimized together with the dynamic compression, see the description below for details.

\subsection{Unified Static and Dynamic Compression}
\begin{figure}[!t]
\centering
\includegraphics[width=0.4\textwidth]{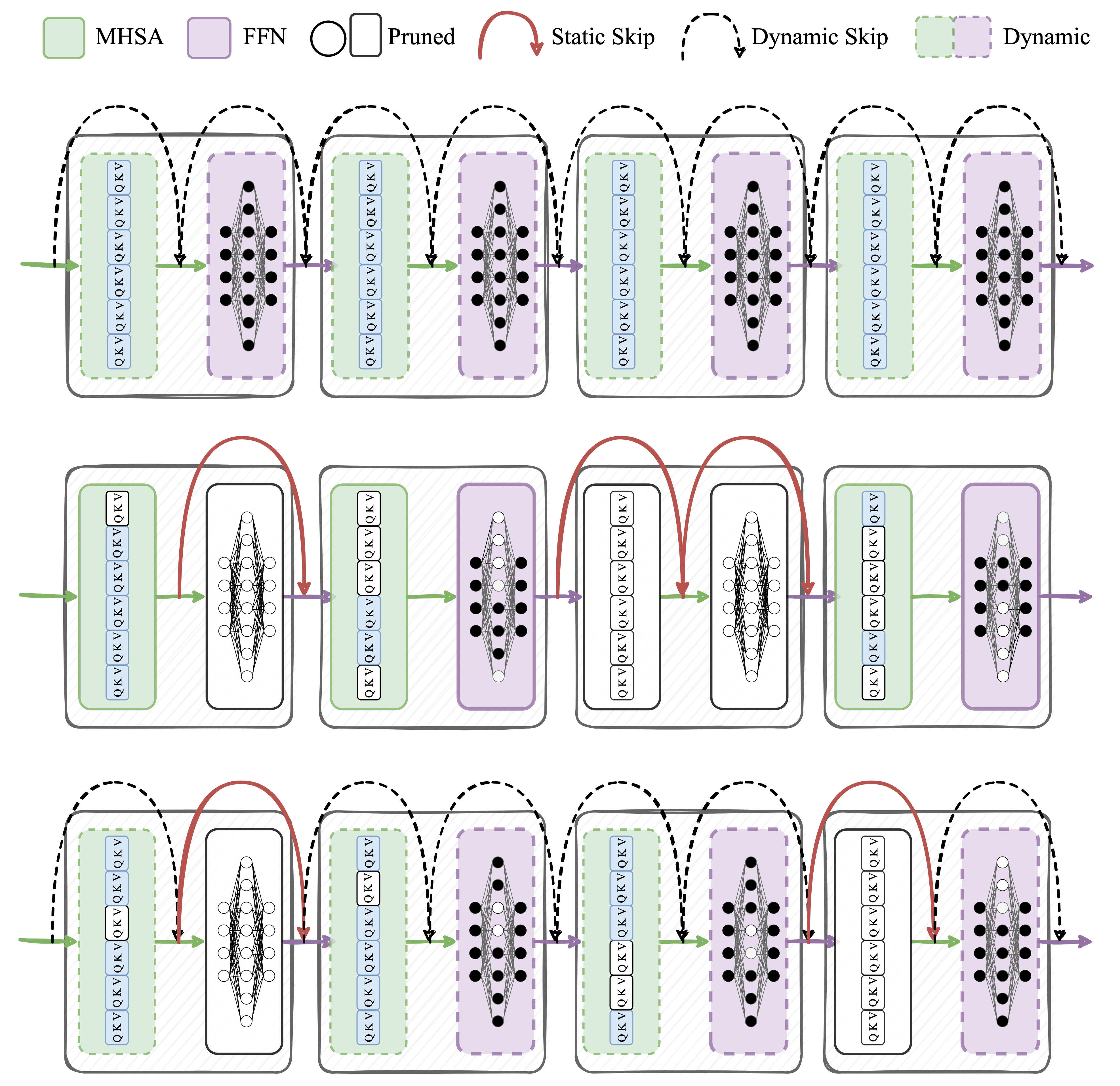}
\caption{An illustration of the optimization results from three types of compression techniques. When reaching the same compression ratio, joint compression will automatically choose optimal dynamic and static percentile schedules to balance the performance and compression ratios.}
\label{fig:illustrationtoy}
\end{figure}

Static compression can explicitly prune the sub-structure of the ViT models, thereby reducing the number of parameters and computation of the ViT models. However, when the retained ratio of the static compression pruned model is too small, it will reduce the representation ability and robustness of the ViT models. Dynamic compression can dynamically execute the network according to the input, but the dynamic decision networks are usually used as the controller, which increases the upper bound of backbone models and inference time. To balance the advantages and shortcomings of these two methods, we propose USDC, a joint compression mechanism that unifies static and dynamic compression together. We illustrate three types of compression techniques in Figure~\ref{fig:illustrationtoy}.

The optimization of USDC is divided into two stages. In the first stage, we use the differentiable neural search method~\cite{liu2018darts} to optimize the static compression parameters to get lightweight ViT models and to search dynamic gates network as described above at Sec.~\ref{sec:dynamic}. We jointly optimize the cross-entropy loss $\mathcal{L}_{cls}$ of the image classification task and compression resource loss $\mathcal{L}_{res}$.  We design the resource loss $\mathcal{L}_{res}$ as the square of the FLOPs difference between the compressing model and source backbone model. As shown in Eq.~\ref{eq:loss}, $\mathcal{F}\left(\boldsymbol{\alpha}; \textbf{g} \right)$ represents the FLOPs of the whole ViT model during the optimization of USDC.
\begin{equation}
\begin{aligned}
\label{eq:loss}
    \mathcal{L}_{total} &=\mathcal{L}_{cls} + \gamma\mathcal{L}_{res} \\
     \mathcal{L}_{res} &= \left(\mathcal{F}\left(\boldsymbol{\alpha}; \textbf{g} \right) - f_t\right)^2
\end{aligned}
\end{equation}
where $f_t$ is the target compression ratio and $\gamma$ is the hyperparameter to balance task loss and resource loss, the value of $\mathcal{F}\left(\boldsymbol{\alpha}; \textbf{g} \right)$ is affected by static compression parameters $\boldsymbol{\alpha}$ and dynamic compression gates $\textbf{g}$ as below:

\begin{equation}
\begin{small}
\begin{aligned}
    &\operatorname{F}_{\text{attn}}^{\prime} =  \boldsymbol{\alpha}_{\ell,0}^a \sum_{h \in H}\boldsymbol{\alpha}_{\ell,i, 0}^h \operatorname{F}_{\text{attn}}\\
    &\operatorname{F}_{\text{ffn}}^{\prime} = \boldsymbol{\alpha}_{\ell,0}^f\sum_{h \in H}\boldsymbol{\alpha}_{\ell,i, 1}^n \operatorname{F}_{\text{ffn}} \\
    &\mathcal{F}\left(\boldsymbol{\alpha}; \textbf{g} \right) = \sum_{l \in {\textbf{L}}} \left(\textbf{g}_{\ell,0}\operatorname{F}_{\text{attn}}^{\prime} + \textbf{g}_{\ell,1}\operatorname{F}_{\text{ffn}}^{\prime} + \sum_{k \in K}\boldsymbol{\alpha}_{\ell,k}^k \operatorname{F}_{\boldsymbol{G},\ell,k}\right) + \text{F}_{o} 
    \label{eq:flops}
\end{aligned}
\end{small}
\end{equation}
Where $\operatorname{F}_{\text{attn}}^{\prime}$ and $ \operatorname{F}_{\text{ffn}}^{\prime}$ are the FLOPs scales of MHSA and FFN block which are normalized by total FLOPs of the model, $\operatorname{F}_{\text{o}}$ is the FLOPs scale of both embedding layers and final classifier layer in ViT model. $\operatorname{F}_{\boldsymbol{G},\ell,:} $ are the FLOPs scale of dynamic gates network space of $\ell_{th}$ encoder layer.

\begin{table*}[!h]
\scriptsize
\centering
\begin{tabular}{lrrccc}
\toprule
\multirow{2}{*}{\textbf{Methods}} &
  \multicolumn{2}{c}{\textbf{Efficiency}} &
  \multicolumn{2}{c}{\textbf{Pruning Options}} &
  \multirow{2}{*}{\textbf{Top-1 ACC.(\%)$\uparrow$}} \\ \cline{2-5}
 &
  \multicolumn{1}{c}{Params.$\downarrow$} &
  \multicolumn{1}{c}{FLOPs$\downarrow$} &
  \multicolumn{1}{l}{Static} &
  \multicolumn{1}{l}{Dynamic} &
   \\ \hline
ViT-base~\cite{yin2022avit}            & 86M            & 17.6G          & - & - & 77.9\%          \\
DeiT-S~\cite{touvron2021deit}               & 22M            & 4.6G           & - & - & 79.8\%          \\
LeViT-256~\cite{Graham_2021levit}              & 18.9M          & 1.1G           & - & - & 80.1\%          \\
T2T-14 ~\cite{yuan2021t2tvit}              & 21.5M          & 4.8G           & - & - & 81.5\%          \\ \midrule
DynamicViT (DeiT-S)~\cite{rao2021dynamicvit}   & 22M            & 3.4G           & - & $\checkmark$ & 78.3\%          \\
A-ViT (DeiT-S)~\cite{yin2022avit}        & 22M            & 3.6G           & - & $\checkmark$ & 78.6\%          \\
MIA-Former (DeiT-S)~\cite{yu2022mia}   & 22M            & 4G             & - & $\checkmark$ & 78.6\%          \\
Evo (LeViT-256)~\cite{xu2022evo}   & 19M            & -             & - & $\checkmark$ & 78.8\%          \\
\textbf{USDC (DeiT-S)} & \textbf{19M}   & \textbf{3.35G} & $\checkmark$ & $\checkmark$ & \textbf{79.0\%} \\
\textbf{USDC (T2T-14)} & \textbf{19.9M} & \textbf{3.4G}  & $\checkmark$ & $\checkmark$ & \textbf{80.6\%} \\
AdaViT (T2T-19)~\cite{meng2022adavit}        & 32M            & 3.9G           & - & $\checkmark$ & 81.1\%          \\
\textbf{USDC (T2T-14)} & \textbf{20.3M} & \textbf{3.7G}  & $\checkmark$ & $\checkmark$ & \textbf{81.2\%} \\ \bottomrule
\end{tabular}
\caption{Our USDC outperforms these recent baseline pruning methods by a clear margin, achieving a better balance between the model efficiency and the compression ratios.}
\label{tab:comparision}
\end{table*}

In the second stage, according to the optimizing results of the static compression parameters $\boldsymbol{\alpha}$ on the first stage, \textbf{we explicitly prune the ViT model and continue to fine-tune the pruned model under dynamic compression constraint}. In the second stage, each transformer encoder layer selects a fixed $\boldsymbol{G}_\ell$ with the largest value in $\boldsymbol{\alpha_{\ell}^g}$. The FLOPs $\mathcal{F}\left(\textbf{g} \right) $ of ViT model on second stage is:
\begin{align}
\small
    &\mathcal{F}\left(\textbf{g} \right) = \sum_{l \in {\textbf{L}}} \left(\textbf{g}_{\ell,0}\operatorname{F}_{\text{attn}}^{\prime\prime}   + \textbf{g}_{\ell,1}\operatorname{F}_{\text{ffn}}^{\prime\prime}  +  \operatorname{F}_{\boldsymbol{G},\ell}\right) + \text{F}_{o} 
\end{align}
Where $\operatorname{F}_{\text{attn}}^{\prime\prime} $ and $\operatorname{F}_{\text{ffn}}^{\prime\prime} $ are the corresponding FLOPs scales after selecting the largest value in $\{\boldsymbol{\alpha}_{\ell}^a, \boldsymbol{\alpha}_{\ell}^m, \boldsymbol{\alpha}_{\ell}^h, \boldsymbol{\alpha}_{\ell}^n\}$ in Eq.~\ref{eq:flops} respectively.
For both stages, dynamic compression is trained with group-level gates, and the $\gamma$ is set to 100.



\section{Experiments}
\paragraph{Datasets and Benchmarks.} We conduct experiments for image classification tasks on ImageNet-1K~\cite{deng2009imagenet} datasets. We use DeiT~\cite{touvron2021deit} and T2T-VIT~\cite{yuan2021t2tvit} as the backbone models due to their strong performance on the image classification task, which both are vision transformer-based models. We compare the Top-1 accuracy and the inference floating point operations (FLOPs) on ImageNet-1K validation dataset.

\paragraph{Main Results}
We compare USDC with several other dynamic compression methods that achieve state-of-the-art results on vision transformer models. DynamicViT~\cite{rao2021dynamicvit} prune redundant tokens via halting decisions relaxed by Gumbel-softmax. A-ViT early learns halting via adaptive computation time~\cite{graves2016adaptive} to stop tokens early. AdaViT~\cite{meng2022adavit} uses three linear layers as the dynamic gate network to control the token, head, and block separately. MIA-Former applies two-layer CNN with one-layer linear as MIA-Controller to dynamic skipping on the head, depth, and token of ViT models. All of those compression methods are dynamic pruning types. Most of the previous dynamic methods on ViT models need to apply additional parameters for the dynamic decision network, which increases the overall number of parameters than the original ViT models~\cite{meng2022adavit, rao2021dynamicvit,yu2022mia}.

The comparison with previous work is summarized in Tab.~\ref{tab:comparision}. To the best of our knowledge, USDC is the first to unify the static and dynamic compression together, the reduction of parameters under static dynamic compression can offset the added parameters on the dynamic gates network. From Tab.~\ref{tab:comparision}, we can notice that all USDC models can reduce the overall parameter compared with backbone models and the other dynamic network method. In comparison, USDC can cut down parameters and FLOPs while achieving the best performance of \textbf{79.0\%}  on DeiT-small compared with DynamicViT, A-ViT, and MIA-Former. \textbf{More experiments and ablations are in the appendix.}

\section{Conclusions}

In this paper, we propose a unified static and dynamic compression framework USDC for the vision transformer, which can be jointly optimized in an end-to-end manner. In the USDC framework, the gate network for each encoder layer will be automatically selected by the NAS for a better trade-off between performance and efficiency. We also proposed the group-level gate augmentation strategy and introduced it into the training for the performance consistency between training and inference. The comparisons with the state-of-the-art compression methods show the effectiveness of USDC. Additionally, the ablation studies show that the group-level strategy and gate network search can benefit the performance of compression. However, our training procedures are still complicated, in the future, we will further explore more efficient and effective unified dynamic and static compression framework designs.

\clearpage
{\small
\bibliography{references}

\begin{thebibliography}{56}
\providecommand{\natexlab}[1]{#1}
\providecommand{\url}[1]{\texttt{#1}}
\expandafter\ifx\csname urlstyle\endcsname\relax
  \providecommand{\doi}[1]{doi: #1}\else
  \providecommand{\doi}{doi: \begingroup \urlstyle{rm}\Url}\fi

\bibitem[Vaswani et~al.(2017)Vaswani, Shazeer, Parmar, Uszkoreit, Jones, Gomez,
  Kaiser, and Polosukhin]{vaswani2017attention}
Ashish Vaswani, Noam Shazeer, Niki Parmar, Jakob Uszkoreit, Llion Jones,
  Aidan~N Gomez, Lukasz Kaiser, and Illia Polosukhin.
\newblock Attention is all you need.
\newblock \emph{arXiv preprint arXiv:1706.03762}, 2017.

\bibitem[Yin et~al.(2022)Yin, Vahdat, Alvarez, Mallya, Kautz, and
  Molchanov]{yin2022avit}
Hongxu Yin, Arash Vahdat, Jose~M Alvarez, Arun Mallya, Jan Kautz, and Pavlo
  Molchanov.
\newblock A-vit: Adaptive tokens for efficient vision transformer.
\newblock In \emph{Proceedings of the IEEE/CVF Conference on Computer Vision
  and Pattern Recognition}, pages 10809--10818, 2022.

\bibitem[Devlin et~al.(2018)Devlin, Chang, Lee, and Toutanova]{devlin2018bert}
Jacob Devlin, Ming-Wei Chang, Kenton Lee, and Kristina Toutanova.
\newblock Bert: Pre-training of deep bidirectional transformers for language
  understanding.
\newblock \emph{arXiv preprint arXiv:1810.04805}, 2018.

\bibitem[Liu et~al.(2021{\natexlab{a}})Liu, Lin, Cao, Hu, Wei, Zhang, Lin, and
  Guo]{liu2021swin}
Ze~Liu, Yutong Lin, Yue Cao, Han Hu, Yixuan Wei, Zheng Zhang, Stephen Lin, and
  Baining Guo.
\newblock Swin transformer: Hierarchical vision transformer using shifted
  windows.
\newblock In \emph{Proceedings of the IEEE/CVF International Conference on
  Computer Vision}, pages 10012--10022, 2021{\natexlab{a}}.

\bibitem[Yuan et~al.(2021)Yuan, Chen, Wang, Yu, Shi, Jiang, Tay, Feng, and
  Yan]{yuan2021t2tvit}
Li~Yuan, Yunpeng Chen, Tao Wang, Weihao Yu, Yujun Shi, Zi-Hang Jiang,
  Francis~EH Tay, Jiashi Feng, and Shuicheng Yan.
\newblock Tokens-to-token vit: Training vision transformers from scratch on
  imagenet.
\newblock In \emph{Proceedings of the IEEE/CVF International Conference on
  Computer Vision}, pages 558--567, 2021.

\bibitem[Chen et~al.(2021)Chen, Cheng, Gan, Yuan, Zhang, and
  Wang]{chen2021chasing}
Tianlong Chen, Yu~Cheng, Zhe Gan, Lu~Yuan, Lei Zhang, and Zhangyang Wang.
\newblock Chasing sparsity in vision transformers: An end-to-end exploration.
\newblock \emph{Advances in Neural Information Processing Systems},
  34:\penalty0 19974--19988, 2021.

\bibitem[Tang et~al.(2021)Tang, Han, Wang, Xu, Guo, Xu, and Tao]{tang2021patch}
Yehui Tang, Kai Han, Yunhe Wang, Chang Xu, Jianyuan Guo, Chao Xu, and Dacheng
  Tao.
\newblock Patch slimming for efficient vision transformers.
\newblock \emph{arXiv preprint arXiv:2106.02852}, 2021.

\bibitem[Yu et~al.(2021)Yu, Chen, Shen, Yuan, Tan, Yang, Liu, and
  Wang]{yu2021uvc}
Shixing Yu, Tianlong Chen, Jiayi Shen, Huan Yuan, Jianchao Tan, Sen Yang,
  Ji~Liu, and Zhangyang Wang.
\newblock Unified visual transformer compression.
\newblock In \emph{International Conference on Learning Representations}, 2021.

\bibitem[Hou and Kung(2022)]{hou2022multi}
Zejiang Hou and Sun-Yuan Kung.
\newblock Multi-dimensional vision transformer compression via dependency
  guided gaussian process search.
\newblock In \emph{Proceedings of the IEEE/CVF Conference on Computer Vision
  and Pattern Recognition}, pages 3669--3678, 2022.

\bibitem[Pan et~al.(2021)Pan, Panda, Jiang, Wang, Feris, and Oliva]{pan2021ia}
Bowen Pan, Rameswar Panda, Yifan Jiang, Zhangyang Wang, Rogerio Feris, and Aude
  Oliva.
\newblock Ia-red2: Interpretability-aware redundancy reduction for vision
  transformers.
\newblock \emph{Advances in Neural Information Processing Systems},
  34:\penalty0 24898--24911, 2021.

\bibitem[Rao et~al.(2021)Rao, Zhao, Liu, Lu, Zhou, and
  Hsieh]{rao2021dynamicvit}
Yongming Rao, Wenliang Zhao, Benlin Liu, Jiwen Lu, Jie Zhou, and Cho-Jui Hsieh.
\newblock Dynamicvit: Efficient vision transformers with dynamic token
  sparsification.
\newblock \emph{Advances in neural information processing systems},
  34:\penalty0 13937--13949, 2021.

\bibitem[Xu et~al.(2022)Xu, Zhang, Zhang, Sheng, Li, Dong, Zhang, Xu, and
  Sun]{xu2022evo}
Yifan Xu, Zhijie Zhang, Mengdan Zhang, Kekai Sheng, Ke~Li, Weiming Dong, Liqing
  Zhang, Changsheng Xu, and Xing Sun.
\newblock Evo-vit: Slow-fast token evolution for dynamic vision transformer.
\newblock In \emph{Proceedings of the AAAI Conference on Artificial
  Intelligence}, volume~36, pages 2964--2972, 2022.

\bibitem[Meng et~al.(2022)Meng, Li, Chen, Lan, Wu, Jiang, and
  Lim]{meng2022adavit}
Lingchen Meng, Hengduo Li, Bor-Chun Chen, Shiyi Lan, Zuxuan Wu, Yu-Gang Jiang,
  and Ser-Nam Lim.
\newblock Adavit: Adaptive vision transformers for efficient image recognition.
\newblock In \emph{Proceedings of the IEEE/CVF Conference on Computer Vision
  and Pattern Recognition}, pages 12309--12318, 2022.

\bibitem[Yu et~al.(2022)Yu, Fu, Li, Li, and Lin]{yu2022mia}
Zhongzhi Yu, Yonggan Fu, Sicheng Li, Chaojian Li, and Yingyan Lin.
\newblock Mia-former: Efficient and robust vision transformers via
  multi-grained input-adaptation.
\newblock In \emph{Proceedings of the AAAI Conference on Artificial
  Intelligence}, volume~36, pages 8962--8970, 2022.

\bibitem[Han et~al.(2021)Han, Huang, Song, Yang, Wang, and
  Wang]{han2021dynamicsurvey}
Yizeng Han, Gao Huang, Shiji Song, Le~Yang, Honghui Wang, and Yulin Wang.
\newblock Dynamic neural networks: A survey.
\newblock \emph{IEEE Transactions on Pattern Analysis and Machine
  Intelligence}, 2021.

\bibitem[Yang et~al.(2020)Yang, Han, Chen, Song, Dai, and
  Huang]{yang2020resolution}
Le~Yang, Yizeng Han, Xi~Chen, Shiji Song, Jifeng Dai, and Gao Huang.
\newblock Resolution adaptive networks for efficient inference.
\newblock In \emph{Proceedings of the IEEE/CVF conference on computer vision
  and pattern recognition}, pages 2369--2378, 2020.

\bibitem[Guan et~al.(2017)Guan, Liu, Liu, and Peng]{guan2017energy}
Jiaqi Guan, Yang Liu, Qiang Liu, and Jian Peng.
\newblock Energy-efficient amortized inference with cascaded deep classifiers.
\newblock \emph{arXiv preprint arXiv:1710.03368}, 2017.

\bibitem[Touvron et~al.(2021{\natexlab{a}})Touvron, Cord, Douze, Massa,
  Sablayrolles, and J{\'e}gou]{touvron2021deit}
Hugo Touvron, Matthieu Cord, Matthijs Douze, Francisco Massa, Alexandre
  Sablayrolles, and Herv{\'e} J{\'e}gou.
\newblock Training data-efficient image transformers \& distillation through
  attention.
\newblock In \emph{International Conference on Machine Learning}, pages
  10347--10357. PMLR, 2021{\natexlab{a}}.

\bibitem[Deng et~al.(2009)Deng, Dong, Socher, Li, Li, and
  Fei-Fei]{deng2009imagenet}
Jia Deng, Wei Dong, Richard Socher, Li-Jia Li, Kai Li, and Li~Fei-Fei.
\newblock Imagenet: A large-scale hierarchical image database.
\newblock In \emph{2009 IEEE conference on computer vision and pattern
  recognition}, pages 248--255. Ieee, 2009.

\bibitem[Touvron et~al.(2021{\natexlab{b}})Touvron, Cord, Sablayrolles,
  Synnaeve, and J{\'e}gou]{touvron2021ceit}
Hugo Touvron, Matthieu Cord, Alexandre Sablayrolles, Gabriel Synnaeve, and
  Herv{\'e} J{\'e}gou.
\newblock Going deeper with image transformers.
\newblock In \emph{Proceedings of the IEEE/CVF International Conference on
  Computer Vision}, pages 32--42, 2021{\natexlab{b}}.

\bibitem[Wang et~al.(2021{\natexlab{a}})Wang, Zhu, Adam, Yuille, and
  Chen]{wang2021max}
Huiyu Wang, Yukun Zhu, Hartwig Adam, Alan Yuille, and Liang-Chieh Chen.
\newblock Max-deeplab: End-to-end panoptic segmentation with mask transformers.
\newblock In \emph{Proceedings of the IEEE/CVF conference on computer vision
  and pattern recognition}, pages 5463--5474, 2021{\natexlab{a}}.

\bibitem[Valanarasu et~al.(2021)Valanarasu, Oza, Hacihaliloglu, and
  Patel]{valanarasu2021medical}
Jeya Maria~Jose Valanarasu, Poojan Oza, Ilker Hacihaliloglu, and Vishal~M
  Patel.
\newblock Medical transformer: Gated axial-attention for medical image
  segmentation.
\newblock In \emph{International Conference on Medical Image Computing and
  Computer-Assisted Intervention}, pages 36--46. Springer, 2021.

\bibitem[Carion et~al.(2020)Carion, Massa, Synnaeve, Usunier, Kirillov, and
  Zagoruyko]{carion2020end}
Nicolas Carion, Francisco Massa, Gabriel Synnaeve, Nicolas Usunier, Alexander
  Kirillov, and Sergey Zagoruyko.
\newblock End-to-end object detection with transformers.
\newblock In \emph{European conference on computer vision}, pages 213--229.
  Springer, 2020.

\bibitem[Arnab et~al.(2021)Arnab, Dehghani, Heigold, Sun, Lu{\v{c}}i{\'c}, and
  Schmid]{arnab2021vivit}
Anurag Arnab, Mostafa Dehghani, Georg Heigold, Chen Sun, Mario Lu{\v{c}}i{\'c},
  and Cordelia Schmid.
\newblock Vivit: A video vision transformer.
\newblock In \emph{Proceedings of the IEEE/CVF International Conference on
  Computer Vision}, pages 6836--6846, 2021.

\bibitem[Liu et~al.(2022)Liu, Ning, Cao, Wei, Zhang, Lin, and Hu]{liu2022video}
Ze~Liu, Jia Ning, Yue Cao, Yixuan Wei, Zheng Zhang, Stephen Lin, and Han Hu.
\newblock Video swin transformer.
\newblock In \emph{Proceedings of the IEEE/CVF Conference on Computer Vision
  and Pattern Recognition}, pages 3202--3211, 2022.

\bibitem[Fan et~al.(2021)Fan, Xiong, Mangalam, Li, Yan, Malik, and
  Feichtenhofer]{fan2021multiscale}
Haoqi Fan, Bo~Xiong, Karttikeya Mangalam, Yanghao Li, Zhicheng Yan, Jitendra
  Malik, and Christoph Feichtenhofer.
\newblock Multiscale vision transformers.
\newblock In \emph{Proceedings of the IEEE/CVF International Conference on
  Computer Vision}, pages 6824--6835, 2021.

\bibitem[Wang et~al.(2018)Wang, Yu, Dou, Darrell, and
  Gonzalez]{wang2018skipnet}
Xin Wang, Fisher Yu, Zi-Yi Dou, Trevor Darrell, and Joseph~E Gonzalez.
\newblock Skipnet: Learning dynamic routing in convolutional networks.
\newblock In \emph{Proceedings of the European Conference on Computer Vision
  (ECCV)}, pages 409--424, 2018.

\bibitem[Guo et~al.(2019)Guo, Yu, Wu, Liang, Qin, and Yan]{guo2019dynamic}
Qiushan Guo, Zhipeng Yu, Yichao Wu, Ding Liang, Haoyu Qin, and Junjie Yan.
\newblock Dynamic recursive neural network.
\newblock In \emph{Proceedings of the IEEE/CVF Conference on Computer Vision
  and Pattern Recognition}, pages 5147--5156, 2019.

\bibitem[Kaya et~al.(2019)Kaya, Hong, and Dumitras]{kaya2019shallow}
Yigitcan Kaya, Sanghyun Hong, and Tudor Dumitras.
\newblock Shallow-deep networks: Understanding and mitigating network
  overthinking.
\newblock In \emph{International conference on machine learning}, pages
  3301--3310. PMLR, 2019.

\bibitem[Bolukbasi et~al.(2017)Bolukbasi, Wang, Dekel, and
  Saligrama]{bolukbasi2017adaptive}
Tolga Bolukbasi, Joseph Wang, Ofer Dekel, and Venkatesh Saligrama.
\newblock Adaptive neural networks for efficient inference.
\newblock In \emph{International Conference on Machine Learning}, pages
  527--536. PMLR, 2017.

\bibitem[Jie et~al.(2019)Jie, Sun, Li, Feng, and Liu]{jie2019anytime}
Zequn Jie, Peng Sun, Xin Li, Jiashi Feng, and Wei Liu.
\newblock Anytime recognition with routing convolutional networks.
\newblock \emph{IEEE transactions on pattern analysis and machine
  intelligence}, 43\penalty0 (6):\penalty0 1875--1886, 2019.

\bibitem[Figurnov et~al.(2017)Figurnov, Collins, Zhu, Zhang, Huang, Vetrov, and
  Salakhutdinov]{figurnov2017spatially}
Michael Figurnov, Maxwell~D Collins, Yukun Zhu, Li~Zhang, Jonathan Huang,
  Dmitry Vetrov, and Ruslan Salakhutdinov.
\newblock Spatially adaptive computation time for residual networks.
\newblock In \emph{Proceedings of the IEEE conference on computer vision and
  pattern recognition}, pages 1039--1048, 2017.

\bibitem[He et~al.(2016)He, Zhang, Ren, and Sun]{he2016deep}
Kaiming He, Xiangyu Zhang, Shaoqing Ren, and Jian Sun.
\newblock Deep residual learning for image recognition.
\newblock In \emph{Proceedings of the IEEE conference on computer vision and
  pattern recognition}, pages 770--778, 2016.

\bibitem[Veit and Belongie(2018)]{veit2018convolutional}
Andreas Veit and Serge Belongie.
\newblock Convolutional networks with adaptive inference graphs.
\newblock In \emph{Proceedings of the European Conference on Computer Vision
  (ECCV)}, pages 3--18, 2018.

\bibitem[Ioffe and Szegedy(2015)]{ioffe2015batch}
Sergey Ioffe and Christian Szegedy.
\newblock Batch normalization: Accelerating deep network training by reducing
  internal covariate shift.
\newblock In \emph{International conference on machine learning}, pages
  448--456. PMLR, 2015.

\bibitem[Wang et~al.(2021{\natexlab{b}})Wang, Huang, Song, Huang, and
  Huang]{wang2021not}
Yulin Wang, Rui Huang, Shiji Song, Zeyi Huang, and Gao Huang.
\newblock Not all images are worth 16x16 words: Dynamic transformers for
  efficient image recognition.
\newblock \emph{Advances in Neural Information Processing Systems},
  34:\penalty0 11960--11973, 2021{\natexlab{b}}.

\bibitem[Guo et~al.(2021)Guo, Yuan, Tan, Wang, Yang, and Liu]{guo2021gdp}
Yi~Guo, Huan Yuan, Jianchao Tan, Zhangyang Wang, Sen Yang, and Ji~Liu.
\newblock Gdp: Stabilized neural network pruning via gates with differentiable
  polarization.
\newblock In \emph{Proceedings of the IEEE/CVF International Conference on
  Computer Vision}, pages 5239--5250, 2021.

\bibitem[Shen et~al.(2020)Shen, Wang, Gui, Tan, Wang, and Liu]{shen2020umec}
Jiayi Shen, Haotao Wang, Shupeng Gui, Jianchao Tan, Zhangyang Wang, and Ji~Liu.
\newblock Umec: Unified model and embedding compression for efficient
  recommendation systems.
\newblock In \emph{International Conference on Learning Representations}, 2020.

\bibitem[Xiao et~al.(2019)Xiao, Wang, and Rajasekaran]{xiao2019autoprune}
Xia Xiao, Zigeng Wang, and Sanguthevar Rajasekaran.
\newblock Autoprune: Automatic network pruning by regularizing auxiliary
  parameters.
\newblock \emph{Advances in neural information processing systems}, 32, 2019.

\bibitem[Lee et~al.(2018)Lee, Ajanthan, and Torr]{lee2018snip}
Namhoon Lee, Thalaiyasingam Ajanthan, and Philip~HS Torr.
\newblock Snip: Single-shot network pruning based on connection sensitivity.
\newblock \emph{arXiv preprint arXiv:1810.02340}, 2018.

\bibitem[Lin et~al.(2018)Lin, Ji, Li, Wu, Huang, and
  Zhang]{lin2018accelerating}
Shaohui Lin, Rongrong Ji, Yuchao Li, Yongjian Wu, Feiyue Huang, and Baochang
  Zhang.
\newblock Accelerating convolutional networks via global \& dynamic filter
  pruning.
\newblock In \emph{IJCAI}, pages 2425--2432, 2018.

\bibitem[Han et~al.(2015)Han, Pool, Tran, and Dally]{han2015learning}
Song Han, Jeff Pool, John Tran, and William Dally.
\newblock Learning both weights and connections for efficient neural network.
\newblock In \emph{Advances in neural information processing systems}, pages
  1135--1143, 2015.

\bibitem[He et~al.(2019)He, Liu, Wang, Hu, and Yang]{he2019filter}
Yang He, Ping Liu, Ziwei Wang, Zhilan Hu, and Yi~Yang.
\newblock Filter pruning via geometric median for deep convolutional neural
  networks acceleration.
\newblock In \emph{Proceedings of the IEEE Conference on Computer Vision and
  Pattern Recognition}, pages 4340--4349, 2019.

\bibitem[Molchanov et~al.(2019)Molchanov, Mallya, Tyree, Frosio, and
  Kautz]{molchanov2019importance}
Pavlo Molchanov, Arun Mallya, Stephen Tyree, Iuri Frosio, and Jan Kautz.
\newblock Importance estimation for neural network pruning.
\newblock In \emph{Proceedings of the IEEE/CVF Conference on Computer Vision
  and Pattern Recognition}, pages 11264--11272, 2019.

\bibitem[Ding et~al.(2021)Ding, Hao, Tan, Liu, Han, Guo, and
  Ding]{Ding_2021_ICCV}
Xiaohan Ding, Tianxiang Hao, Jianchao Tan, Ji~Liu, Jungong Han, Yuchen Guo, and
  Guiguang Ding.
\newblock Resrep: Lossless cnn pruning via decoupling remembering and
  forgetting.
\newblock In \emph{Proceedings of the IEEE/CVF International Conference on
  Computer Vision (ICCV)}, pages 4510--4520, October 2021.

\bibitem[He et~al.(2017)He, Zhang, and Sun]{he2017channel}
Yihui He, Xiangyu Zhang, and Jian Sun.
\newblock Channel pruning for accelerating very deep neural networks.
\newblock In \emph{The IEEE International Conference on Computer Vision
  (ICCV)}, Oct 2017.

\bibitem[Liu et~al.(2017)Liu, Li, Shen, Huang, Yan, and Zhang]{Liu2017learning}
Zhuang Liu, Jianguo Li, Zhiqiang Shen, Gao Huang, Shoumeng Yan, and Changshui
  Zhang.
\newblock Learning efficient convolutional networks through network slimming.
\newblock In \emph{ICCV}, 2017.

\bibitem[Zhu et~al.(2021)Zhu, Han, Tang, and Wang]{zhu2021visual}
Mingjian Zhu, Kai Han, Yehui Tang, and Yunhe Wang.
\newblock Visual transformer pruning.
\newblock \emph{arXiv preprint arXiv:2104.08500}, 2021.

\bibitem[Liu et~al.(2021{\natexlab{b}})Liu, Wang, Han, Ma, and
  Gao]{liu2021posttraining}
Zhenhua Liu, Yunhe Wang, Kai Han, Siwei Ma, and Wen Gao.
\newblock Post-training quantization for vision transformer,
  2021{\natexlab{b}}.

\bibitem[Dosovitskiy et~al.(2020)Dosovitskiy, Beyer, Kolesnikov, Weissenborn,
  Zhai, Unterthiner, Dehghani, Minderer, Heigold, Gelly,
  et~al.]{dosovitskiy2020vit}
Alexey Dosovitskiy, Lucas Beyer, Alexander Kolesnikov, Dirk Weissenborn,
  Xiaohua Zhai, Thomas Unterthiner, Mostafa Dehghani, Matthias Minderer, Georg
  Heigold, Sylvain Gelly, et~al.
\newblock An image is worth 16x16 words: Transformers for image recognition at
  scale.
\newblock \emph{arXiv preprint arXiv:2010.11929}, 2020.

\bibitem[Ba et~al.(2016)Ba, Kiros, and Hinton]{ba2016layernorm}
Jimmy~Lei Ba, Jamie~Ryan Kiros, and Geoffrey~E Hinton.
\newblock Layer normalization.
\newblock \emph{arXiv preprint arXiv:1607.06450}, 2016.

\bibitem[Liu et~al.(2018)Liu, Simonyan, and Yang]{liu2018darts}
Hanxiao Liu, Karen Simonyan, and Yiming Yang.
\newblock Darts: Differentiable architecture search.
\newblock \emph{arXiv preprint arXiv:1806.09055}, 2018.

\bibitem[Maddison et~al.(2016)Maddison, Mnih, and Teh]{maddison2016gumbel}
Chris~J Maddison, Andriy Mnih, and Yee~Whye Teh.
\newblock The concrete distribution: A continuous relaxation of discrete random
  variables.
\newblock \emph{arXiv preprint arXiv:1611.00712}, 2016.

\bibitem[Graham et~al.(2021)Graham, El-Nouby, Touvron, Stock, Joulin, J\'egou,
  and Douze]{Graham_2021levit}
Benjamin Graham, Alaaeldin El-Nouby, Hugo Touvron, Pierre Stock, Armand Joulin,
  Herv\'e J\'egou, and Matthijs Douze.
\newblock Levit: A vision transformer in convnet's clothing for faster
  inference.
\newblock In \emph{Proceedings of the IEEE/CVF International Conference on
  Computer Vision (ICCV)}, pages 12259--12269, October 2021.

\bibitem[Graves(2016)]{graves2016adaptive}
Alex Graves.
\newblock Adaptive computation time for recurrent neural networks.
\newblock \emph{arXiv preprint arXiv:1603.08983}, 2016.

\bibitem[Loshchilov and Hutter(2017)]{loshchilov2017decoupled}
Ilya Loshchilov and Frank Hutter.
\newblock Decoupled weight decay regularization.
\newblock \emph{arXiv preprint arXiv:1711.05101}, 2017.

\end{thebibliography}
}

\appendix

\section{Implementation Details}
USDC experiments are conducted on DeiT-small and T2T-Vit-14 models. The DeiT-small has $L = 12$ transformer encoder layers with $H=6$ heads in the MHSA block.  The T2T-Vit-14 model has $L = 14$ encoder layers with $H=7$ heads in the MHSA block. We initialize the ViT backbones of USDC with the pre-trained parameters released in the official reports~\cite{touvron2021deit, yuan2021t2tvit}.
We implement the USDC compression mechanism on all transformer layers of ViT models including the first encoder layer.

USDC code is implemented with PyTorch, and all experiments are conducted on 8 NVIDIA V100 GPUs. For all experiments, we trained the models with the learning rate of $5e^{-4}$, weight decay of 0.05 for 300 epochs, and a training batch size is 256. AdamW~\cite{loshchilov2017decoupled} optimizer is used with a cosine learning rate schedule. The input size is set to $224\times 224$. The temperature of Gumbel Softmax is set to $5.0$ for dynamic skipping of USDC, and $2.0$ for dynamic gates networks searching and static compression in USDC. 

\section{Ablation Studies}

\begin{figure}[!h]
\centering
\includegraphics[width=0.4\textwidth]{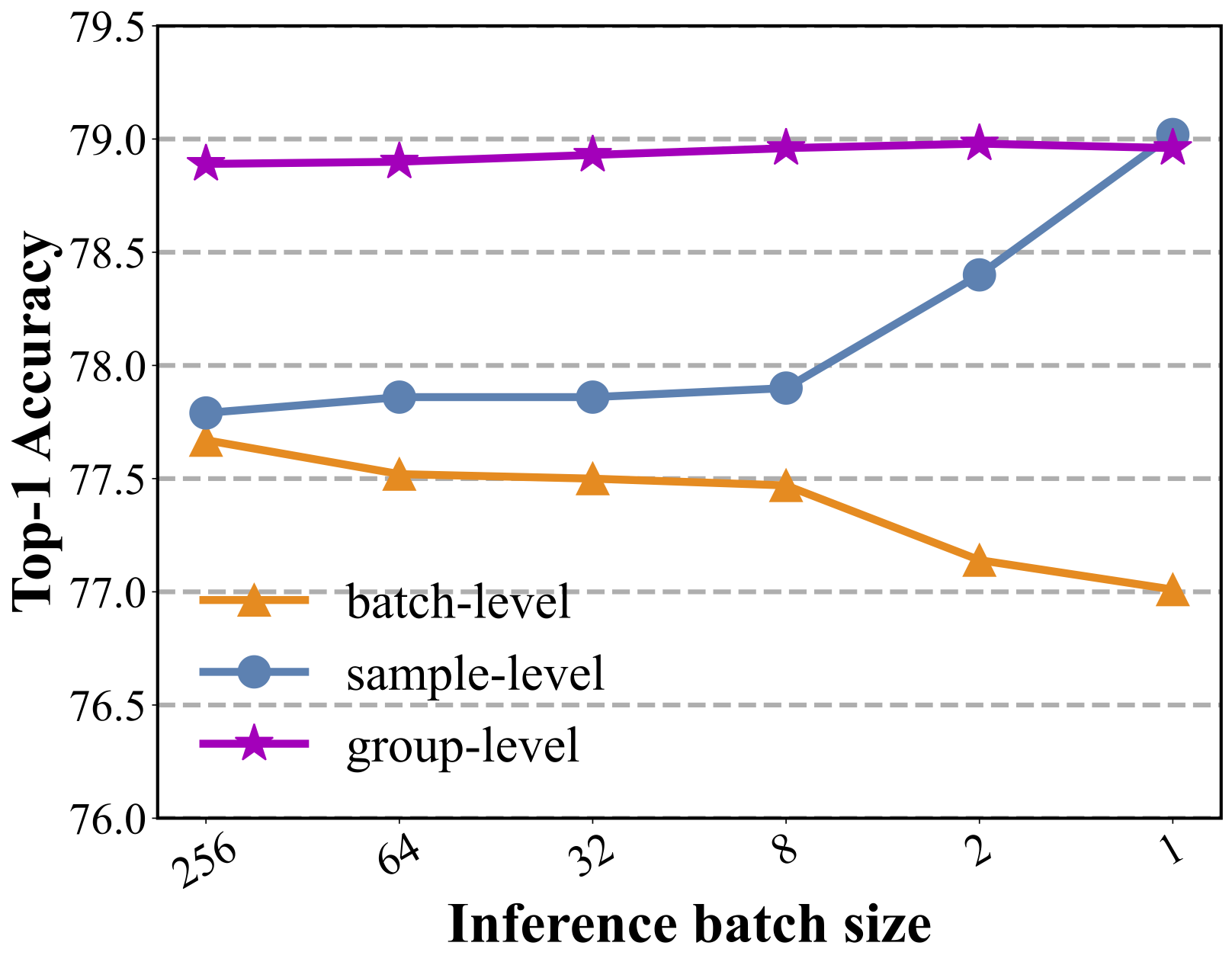}
\caption{We demonstrate the different performances for different inference batch sizes, while the training batch size is always 256. Our proposed group-level gate augmentation can consistently maintain good performance by balancing the influence of batch-level and sample-level gates.}
\label{fig:batch}
\end{figure}

\paragraph{Comparisons of different levels of dynamic gates.}  We compare the performance of USDC on three computing gate levels. Sample level calculates a unique gate for each sample in a batch during training. Batch-level calculates the mean of gate features at batch dimension and then gets only one gate for the whole batch.  Group-level split batch by sub-groups and calculate gate under the sub-group, which can simulate various batch sizes and balance the deviation caused by the inconsistency of batch size between train and inference. We trained the USDC method by setting batch-level, sample-level, and group-level gates on DeiT-Small separately, and the training batch size is $256$. In actual deployment, the inference batch size is usually smaller than the train batch size, therefore,  we evaluate the performance of those three gate levels on several different inference batch sizes from  $256$ to $1$. As shown in Fig~\ref{fig:batch}, we notice that batch-level has more accuracy drop after compression. Sample-level achieves better evaluation performance at the smallest inference batch size $1$, but there is an accuracy decrease when the inference batch becomes bigger, the maximum decrease of sample-level on different inference batch sizes is $\Delta 1.23\%$. The accuracy difference of group levels between different batch sizes is only $\Delta 0.09\%$. 

\paragraph{Comparisons of different dynamic decision networks.} We compare the performance of USDC which dynamic decision networks obtained by automated search method with the manually selected way. The searching method selects a unique dynamic decision network for each block. For manual selection, we selected dynamic decision networks with the largest parameters and smallest parameters in the search space for ablation experiments. As shown in Tab.~\ref{tab:gatenetwoek}, the Top-1 accuracy on Imagenet-1K of searched dynamic decision networks is better than manually selected.

\paragraph{Comparisons of different pruning options.}  In Tab.~\ref{tab:3}, we regard USDC as a combined static and dynamic compression method, and we compare USDC with only static compression methods and only dynamic compression methods on the DeiT-S model. It can be noticed that the joint static and dynamic method in USDC has achieved the highest effect. A similar illustration is also shown in Fig.~\ref{fig:illustration}.

\begin{table}
\small
\centering
\begin{tabular}{c|cccc}
\toprule
\multirow{2}*{\textbf{Effects of $\boldsymbol{G}$} }  & \multirow{2}*{Batch}  &  \multicolumn{3}{c}{\textbf{Top-1 Accuracy (\%)}$\uparrow$}      \\ 
\cline{3-5}
                     &                      &    $\textbf{Manual}$       &     $\textbf{Manual}$      & $\textbf{Search}$\\ \midrule
\multirow{3}{*}{\begin{tabular}[c]{@{}c@{}}\textbf{USDC (DeiT-S)} \end{tabular}} 
                & 256    & 78.44 & 78.32 & \textbf{78.89}   \\
               & 32 & 78.44 & 78.30 & \textbf{78.93} \\
               & 2  & 78.46 & 78.30 & \textbf{78.98} \\ \midrule
FLOPs of $\boldsymbol{G}_*$   & -  & 0.89M & \textbf{0.02M} & 0.75M          \\
FLOPs of model & -  & 3.37G & 3.4G  & \textbf{3.35G} \\ \bottomrule
\end{tabular}
\caption{\textbf{Effects of different dynamic decision networks.} Batch means the inference batch size during evaluation. The first column manually chooses $\boldsymbol{G}_{:,0}$ as a dynamic decision network for all encoder layers.  And the second column manually chooses $\boldsymbol{G}_{:,2}$ for all layers. The third column automatically searches dynamic decision networks as described at Seq.~\ref{sec:dynamic}.}
\label{tab:gatenetwoek}
\end{table}

\begin{table}
\centering
\small
\begin{tabular}{r|ccc}
\toprule
Pruning Options & Static  & Dynamic & Static \& Dynamic \\ \midrule
Top-1 ACC. & 77.97\% & 78.13\% & \textbf{78.96\%}         \\
FLOPs & 3.40G   & 3.36G   & \textbf{3.35G}           \\ \bottomrule
\end{tabular}
\caption{Comparisons of different pruning options on DeiT-S demonstrate the superiority of our joint optimization.}
\label{tab:3}
\end{table}


\section{More Compression Comparisons}
To further demonstrate the superiority of our method when the compression rate is smaller than $50\%$, we conduct the USDC on the T2T-ViT-19~\cite{yuan2021t2tvit} model. From Tab.~\ref{tab:t2t19}, we can find that USDC reduces the 55.3\% FLOPs of the original baseline T2T-ViT-19 while the drop of Top-1 accuracy on Imagenet-1K~\cite{deng2009imagenet} dataset is only $0.6\%$. Our USDC framework outperforms the previous dynamic compression method AdaViT~\cite{meng2022adavit} with fewer parameters and FLOPs. It should be noticed that AdaViT takes three linear layers as the dynamic decision network for each transformer layer in T2T-ViT-19, thus the number of parameters in AdaViT is larger than the original T2T-ViT-19.

\begin{table}[!h]
\small
\centering
\caption{The comparisons on the T2T-ViT-19 model.}
\begin{tabular}{l|ccc}
\toprule
\textbf{Method}             & \textbf{Params.} & \textbf{FLOPs} & \textbf{Top-1 Acc.}  \\ \hline
Baseline (T2T-ViT-19)~\cite{yuan2021t2tvit}         & 39.2M  & 8.5G  & 81.9\% \\
Ada-ViT (T2T-ViT-19)~\cite{meng2022adavit} & $\textgreater$39.2M  & 3.9G  & 81.1\% \\
\textbf{USDC (T2T-ViT-19)}   & \textbf{34.4M}  & \textbf{3.8G}  & \textbf{81.3\%} \\ \bottomrule
\end{tabular}
\label{tab:t2t19}
\end{table}

\section{Ablations for sub-groups split methods}
Whether the samples within a mini-batch ought to use the same gate controlling at any location of ViT during the training of the dynamic network is a heuristic question.  Different from \textbf{sample-level} strategy~\cite{meng2022adavit,wang2018skipnet,yin2022avit} and \textbf{batch-level} strategy, we propose a \textbf{group-level} strategy to augment dynamic gates. Group-level split batch by sub-groups and calculate gate under the sub-group, which can simulate various smaller batch sizes and mitigate the performance deviation caused by the inconsistency of batch size between train and inference stages. USDC recursively split the gates features into the different groups by a logarithm of 2.

For \textbf{group-level} strategy, we trained the USDC method by splitting sub-groups on average, on random, and our recursive split method separately. The training is on the DeiT-small model on the Imagenet-1K dataset, the training batch size is 256 and all other parameter settings are the same. As shown in Tab.~\ref{tab:group_compare},  we compare different sub-groups split method,  the
The top-1 accuracy on Imagenet-1K of our recursive split method separately is better than the average and random methods.  

\begin{table}[!h]
\small
\centering
\caption{The comparisons of different sub-groups split methods for group-level gate augmentation strategy. The first column splits sub-groups uniformly with step size 32. The second column splits sub-groups uniformly with step size 8. The third column splits the sub-groups randomly with step size ranges in $[1,64]$. The fourth column (Ours) splits sub-groups recursively by a logarithm of 2.}
\begin{tabular}{c|l|cccc}
\toprule
\multirow{2}*{\textbf{Model}}  & \multirow{2}*{$B$}  &  \multicolumn{4}{c}{\textbf{Top-1 Accuracy (\%)}}      \\ 
\cline{3-6}
                     &                      &    Avg-32       &     Avg-8    &     Random     & \textbf{Ours}\\
\midrule
\multirow{6}{*}{\begin{tabular}[c]{@{}c@{}}USDC\\ (DeiT-S)\end{tabular}} & 256 &  77.13 & 77.43 & 77.86 & \textbf{78.89} \\
      & 64 &  77.11 & 77.62 & 77.44 & \textbf{78.90} \\
      
      & 32 &   77.10  & 77.62 & 77.45 & \textbf{78.93} \\
      & 8  &   77.05  & 77.65  & 77.50 & \textbf{78.96} \\
      & 2  &   77.00  & 77.62  & 77.49 & \textbf{78.98} \\
      & 1  &   76.91   & 77.60 & 77.49 & \textbf{78.96}       \\ \midrule
FLOPs & -  &   3.30G   & 3.36G & 3.30G &  3.35G \\ \bottomrule
\end{tabular}
\label{tab:group_compare}
\end{table}

\section{Visualizations}
\begin{figure*}[!b]
\centering
\includegraphics[width=0.6\textwidth]{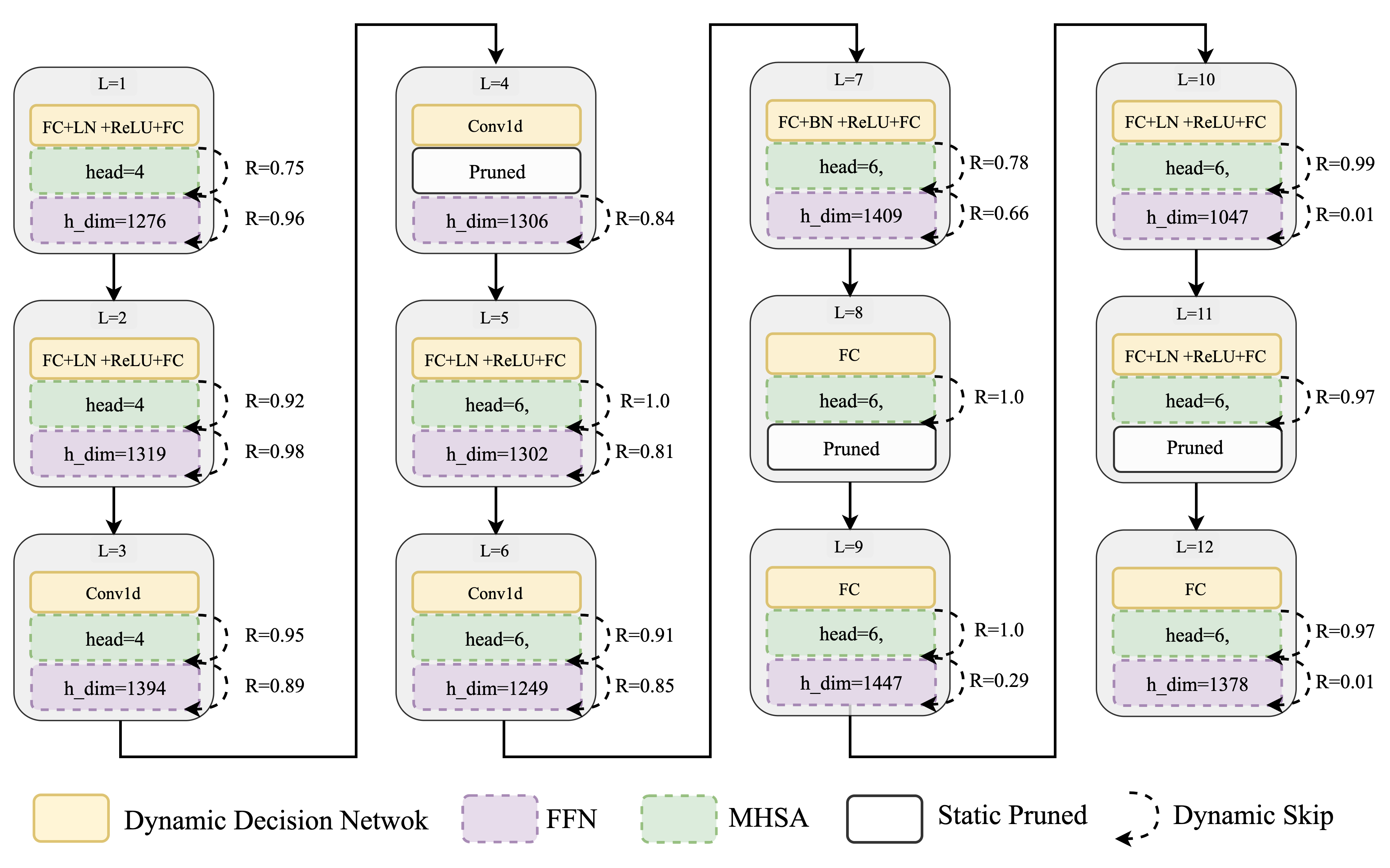}
\caption{An architecture illustration for the compressed DeiT-small model using our USDC. The $L$ is the index of transformer encoder layers. The $head$ is the head number of the MHSA block. The $h\_dim$ is the hidden dimension of FFN blocks. The $R$ is the dynamically executing rate on average for each block. The remaining FLOPs achieved by static compression is 74.9\%. The final remaining FLOPs achieved by joint static and dynamic compression is 64.8\%.}
\label{fig:illustration}
\end{figure*}

We illustrate the structures of the compressed DeiT-Small model by USDC at Fig.~\ref{fig:illustration}. We trained the model in Fig.~\ref{fig:illustration} by unified static and dynamic compression described in the main text. We can notice that the head number of MHSA and the hidden dimension of FFN were reduced by static compression, and some blocks were pruned by static compression of USDC. Meanwhile, the dynamic compression of USDC skipped each block adaptively according to the input features of each transformer layer. The FLOPs of all 12 dynamic decision networks together are only $0.45$M, and the FLOPs of the original  DeiT-small is $4.6$G. As shown in Fig.~\ref{fig:illustration}, the remaining FLOPs achieved by only the static compression part is 74.9\%. The final remaining FLOPs achieved by joint static and dynamic compression is 64.8\%.

\end{document}